\documentclass[runningheads]{llncs}
\usepackage{refcount}
\usepackage{amsmath}
\usepackage{graphicx}
\usepackage{rotating}
\usepackage{multirow}
\usepackage{pgfmath}
\usepackage{pgfplots}
\pgfplotsset{width=4.3cm,compat=1.6}
\usepackage{pgfplotstable}
\usepackage[caption=true]{subfig}
\usepackage{textpos}
\usepackage{enumitem} 
\usepackage[hyphens]{url}
\usepackage[misc]{ifsym}
\begin{document}
\title{OMBA: User-Guided Product Representations for Online Market Basket Analysis\thanks{This research was financially supported by Melbourne Graduate Research Scholarship and Rowden White Scholarship.}}
\titlerunning{Product Representations for Online Market Basket Analysis}
%
\author{Amila Silva \textsuperscript{\Letter} \and
Ling Luo \and
Shanika Karunasekera \and 
Christopher Leckie}
\authorrunning{Amila Silva et al.}
%
\toctitle{OMBA: User-Guided Product Representations for Online Market Basket Analysis}
\tocauthor{Amila~Silva, Ling~Luo, Shanika~Karunasekera, Christopher~Leckie
}

\institute{School of Computing and Information Systems\\ The University of Melbourne \\Parkville, Victoria, Australia\\
\email{\{amila.silva@student., ling.luo@, karus@, caleckie@\}unimelb.edu.au}}
\begin{textblock*}{100mm}(0.05\textwidth,-3.5cm)
\centering \large Accepted as a conference paper at ECML-PKDD 2020
\end{textblock*}
\vspace{-1cm}
{\let\newpage\relax\maketitle}
%
\begin{abstract}
Market Basket Analysis (MBA) is a popular technique to identify associations between products, which is crucial for business decision making. Previous studies typically adopt conventional frequent itemset mining algorithms to perform MBA. However, they generally fail to uncover rarely occurring associations among the products at their most granular level. Also, they have limited ability to capture temporal dynamics in associations between products. Hence, we propose OMBA, a novel representation learning technique for Online Market Basket Analysis. OMBA jointly learns representations for products and users such that they preserve the temporal dynamics of product-to-product and user-to-product associations. Subsequently, OMBA proposes a scalable yet effective online method to generate products' associations using their representations. Our extensive experiments on three real-world datasets show that OMBA outperforms state-of-the-art methods by as much as 21\%, while emphasizing rarely occurring strong associations and effectively capturing temporal changes in associations.
\keywords{Market Basket Analysis \and Online Learning \and Item Representations \and Transaction Data}
\end{abstract}
\section{Introduction}~\label{sec:intro}
\textbf{Motivation.} Market Basket Analysis (MBA) is a technique to uncover relationships (i.e., association rules) between the products that people buy as a basket at each visit. MBA is widely used in today's businesses to gain insights for their business decisions such as product shelving and product merging. For instance, assume there is a strong association (i.e., a higher probability to buy together) between product $p_i$ and product $p_j$. Then both $p_i$ and $p_j$ can be placed on the same shelf to encourage the buyer of one product to buy the other. 

Typically, the interestingness of the association between product $p_i$ and product $p_j$ is measured using \textit{Lift}: the \textit{Support} (i.e., probability of occurrence) of $p_i$ and $p_j$ together divided by the product of the individual \textit{Support} values of $p_i$ and $p_j$ as if they are independent. MBA attempts to uncover the sets of products with high \textit{Lift} scores. However, it is infeasible to compute the \textit{Lift} measure between all product combinations for a large store in today's retail industry, as they offer a broad range of products. For example, Walmart offers more than 45 million products as of 2018\footnote{\url{https://bit.ly/how_many_products_does_walmart_grocery_sell_july_2018}}. Well-known MBA techniques~\cite{agrawal1994fast,deng2014fast,fournier2012mining} can fail to conduct accurate and effective analysis for such a store for the following reasons.

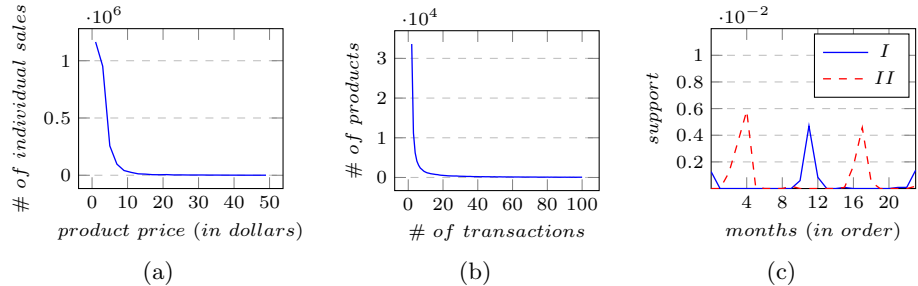
\begin{figure*}[t]
\scriptsize
\centering
\subfloat[]{%
\begin{tikzpicture}
\label{fig:a}
\begin{axis}[
    xlabel={$product\text{ }price\text{ }(in\text{ }dollars)$},
    ylabel={$\#\text{ }of\text{ }individual\text{ }sales$},
    xtick={0, 10, 20, 30, 40, 50},
    legend pos=south east,
    ymajorgrids=true,
    grid style=dashed,
]
\addplot[
    color=blue,
    each nth point={1},
    line width= 0.5pt,
    ] table[x=t,y=v,col sep=comma] {ex4.csv};
\end{axis}
\end{tikzpicture}%
}\hspace{1em}
\subfloat[]{%
\begin{tikzpicture}
\label{fig:b}
\begin{axis}[
    xlabel={$\#\text{ }of\text{ }transactions$},
    ylabel={$\#\text{ }of\text{ }products$},
    xtick={0,20,40,60,80,100},
    legend pos=south east,
    ymajorgrids=true,
    grid style=dashed,
]
\addplot[
    color=blue,
    each nth point={1},
    line width= 0.5pt,
    ] table[x=t,y=v,col sep=comma] {ex5.csv};
\end{axis}
\end{tikzpicture}
}
\hspace{1em}
\subfloat[]{%
\begin{tikzpicture}
\label{fig:c}
\begin{axis}[
    xlabel={$months\text{ }(in\text{ }order)$},
    ylabel={$support$},
    xmin=0, xmax=23,
    ymin=0, ymax=0.012,
    xtick={4,8,12,16,20},
    ytick={0.002,0.004,0.006,0.008,0.01},
    legend pos=north east,
    ymajorgrids=true,
    grid style=dashed,
]
\addplot[
    color=blue,
    each nth point={1},
    line width= 0.5pt,
    ] table[x=t,y=v,col sep=comma] {ex1.csv};
    \addlegendentry{$I$}
    color=blue,
    \addplot[
    color=red,
    dash pattern=on 3pt off 3pt,
    each nth point={1},
    line width= 0.5pt,
    ] table[x=t,y=v,col sep=comma] {ex3.csv};
    \addlegendentry{$II$}
\end{axis}
\end{tikzpicture}
}
\caption{(a) Number of products' sales with respect to the products' prices; (b) number of products with respect to their appearances in the transactions; and (c) temporal changes in the \textit{Support} of
: (I) Valentine Gifts and Decorations; and (II) Rainier Cherries. The plots are generated using Complete Journey dataset}\label{fig:example1}
\end{figure*}

\textbf{Research Gaps.} First, MBA is typically performed using frequent itemset mining algorithms~\cite{deng2014fast,fournier2012mining,han2004mining,agrawal1994fast}
, which produce itemsets whose \textit{Support} is larger than a predefined \textit{minimum Support} value. However, such frequency itemset mining algorithms fail to detect associations among the products with low \textit{Support} values (e.g., expensive products, which are rarely bought, as shown in Figure~\ref{fig:a}), but which are worth analysing. To further elaborate, almost all these algorithms compute \textit{Support} values of itemsets, starting from the smallest sets of size one and gradually increasing the size of the itemsets. If an itemset $P_i$ fails to meet the \textit{minimum Support} value, all the supersets of $P_i$ are pruned from the search space as they cannot have a \textit{Support} larger than the \textit{Support} of $P_i$. Nevertheless, the supersets of $P_i$ can have higher \textit{Lift} scores than $P_i$. Subsequently, the selected itemsets are further filtered to select the itemsets with high \textit{Lift} scores as associated products. 
For instance, \textit{`Mexican Seasoning Mixes'} and \textit{`Tostado Shells'} have $0.012$ and $0.005$ support values in Complete Journey dataset (see Section~\ref{sec:experiments}) 
respectively. Because of the low support of the latter, conventional MBA techniques could fail to check the association between these two products despite them having a strong association with a $Lift$ score of $43.45$. 
Capturing associations of products that have lower \textit{Support} values is important due to the power law distribution of products' sales~\cite{hariharan2013seasonal}, where a huge fraction of
the products have a low sales volume (as depicted in Figure~\ref{fig:b}).  

Second, most existing works perform MBA at a coarser level, which groups multiple products, due to two reasons: (1) data sparsity at the finer levels, which requires a lower \textit{minimum Support} value to capture association rules; and (2) large numbers of different products at the finer levels. Both aforementioned reasons substantially increase the computation time of conventional association rule mining techniques. As a solution to this, several previous works~\cite{ReshuAgarwal2019,Han1995} attempt to initially find strong associations at a coarser level, such as groups of products, and further analyse only the individual products in those groups to identify associations at the finer levels. Such approaches do not cover the whole search space at finer levels, and thus fail to detect the association rules that are only observable at finer levels. In addition, almost all the conventional approaches consider each product as an independent element. This is where representation learning based techniques, which learn a low-dimensional vector for each unit\footnote{``Units'' refers to the attribute values (could be products or users) of the baskets} such that they preserve the semantics of the units, have an advantage. Thus, representation learning techniques are useful to alleviate the cold-start problem (i.e., detecting associations that are unseen in the dataset) and data sparsity. While there are previous works on applying representation learning for shopping basket recommendation, \textit{none of the previous representation learning techniques has been extended to mine association rules between products}. Moreover, user details of the shopping baskets could be useful to understand the patterns of the shopping baskets. Users generally exhibit repetitive buying patterns, which could be exploited by jointly learning representations for users and products in the same embedding space. While such user behaviors have previously been studied in the context of personalized product recommendation~\cite{wan2018representing,le2017basket}, they have not been exploited in the context of association rule mining of products.

Third, conventional MBA techniques consider the set of transactions as a single batch to generate association rules between products, despite that the data may come in a continuous stream. The empirical studies in~\cite{papavasileiou2011time,hariharan2013seasonal} show that the association rules of products deviate over time due to various factors (e.g., seasonal variations and socio-economic factors). To further illustrate this point, Figure~\ref{fig:c} shows the changes in sales of two product categories over time at a retailer. As can be seen, there are significant variations of the products' sales. Consequently, the association rules of such products vary over time. Conventional MBA techniques fail to capture these temporal variations. As a solution to this problem in~\cite{papavasileiou2011time,hariharan2013seasonal}, the transactions are divided into different time-bins and conventional MBA techniques are used to generate association rules for different time bins from scratch. Such approaches are computationally and memory intensive; and ignore the dependencies between consecutive time bins. 

\textbf{Contribution.} In this paper, we propose a novel representation learning based technique to perform Online Market Basket Analysis (\textsc{OMBA}), which: (1) \textit{jointly learns representations for products and users such that the representations preserve their co-occurrences, while emphasizing the products with higher selling prices (typically low in \textit{Support}) and exploiting the semantics of the products and users}; (2) \textit{proposes an efficient approach to perform MBA using the learned product representations, which is capable of capturing rarely occurring and unseen associations among the products}; and (3) \textit{accommodates online updating for product representations, which adapts the model to the temporal changes, without overfitting to the recent information or storing any historical records}. The code for all the algorithms presented in this paper and the data used in the evaluation are publicly available via \url{https://bit.ly/2UwHfr0}.

\section{Related Work}\label{sec:related}

\begin{figure}[t]
    \centering
    \includegraphics[width=\linewidth]{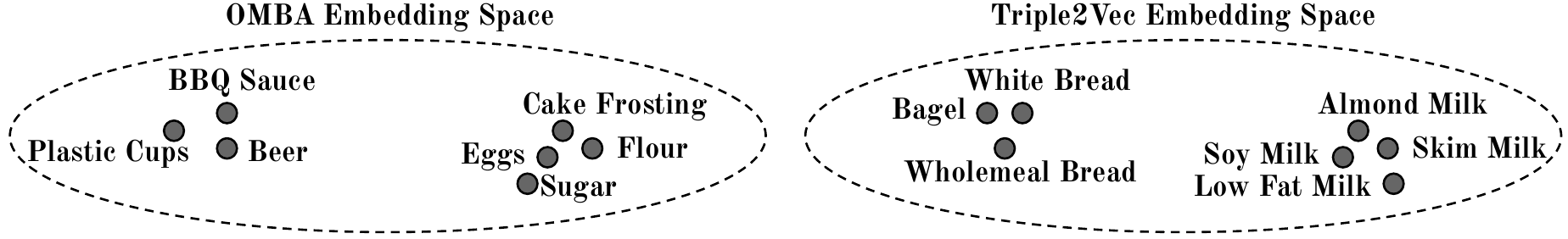}
    \caption{An illustration of the embedding spaces of \textsc{OMBA}, which preserves complementarity, and \textsc{Triple2Vec}, which only preserves the semantic similarity}
    \label{fig:example2}
\end{figure}

\textbf{Conventional MBA Techniques. }Typically, MBA is performed using conventional frequent itemset mining algorithms~\cite{deng2014fast,fournier2012mining,han2004mining,agrawal1994fast}, which initially produce the most frequent (i.e., high \textit{Support}) itemsets in the dataset, out of which the itemsets with high \textit{Lift} scores are subsequently selected as association rules. These techniques mainly differ from each other based on the type of search that they use to explore the space of itemsets (e.g., a depth-first search strategy is used in~\cite{han2004mining}, and the work in~\cite{agrawal1994fast} applies breadth-first search). As elaborated in Section~\ref{sec:intro}, these techniques have the limitations: (1) inability to capture important associations among rarely bought products, which covers a huge fraction of products (see Figure~\ref{fig:b}); (2) inability to alleviate sparsity at the finer product levels; and (3) inability to capture temporal dynamics of association rules. In contrast, OMBA produces association rules from the temporally changing representations of the products to address these limitations.   

\textbf{Representation Learning Techniques. } Due to the importance of capturing the semantics of products, there are previous works that adopt representation learning techniques for products~\cite{wan2018representing,Bai2018,le2017basket,Yu2016,barkan2016item2vec,grbovic2015commerce}. Some of these works address the \textit{next basket recommendation task}, which attempts to predict the next shopping basket of a customer given his/her previous baskets. Most recent works~\cite{Yu2016,Bai2018} in this line adopt recurrent neural networks to model long-term sequential patterns in users' shopping baskets. However, our task differs from these works by concentrating on the MBA task, which models the shopping baskets without considering the order of the items in each. The work proposed in~\cite{le2017basket}, for \textit{product recommendation tasks}, need to store large knowledge graphs (i.e,  co-occurrences matrices), thus they are not ideal to perform MBA using transaction streams. In~\cite{wan2018representing,barkan2016item2vec,grbovic2015commerce}, word2vec language model~\cite{mikolov2013distributed} has been adopted to learn product representations in an online fashion. Out of them, \textsc{Triple2Vec}~\cite{wan2018representing} is the most similar and recent work for our work, which jointly learns representation for products and users by adopting word2vec. However, \textsc{Triple2Vec} learns two embedding spaces for products such that each embedding space preserves semantic similarity (i.e., second-order proximity) as illustrated in Figure~\ref{fig:example2}. Thus, the products' associations cannot be easily discovered from such an embedding space. In contrast, the embedding space of \textsc{OMBA} mainly preserves the complementarity (i.e., first-order proximity). Thus, the associated products are closely mapped in the embedding space as illustrated in Figure~\ref{fig:example2}. Moreover, none of the aforementioned works proposes a method to generate association rules from products' embeddings, which is the ultimate objective of MBA.  

\textbf{Online Learning Techniques. }OMBA learns online representations for products, giving importance to recent information to capture the temporal dynamics of products' associations. 
However, it is challenging to incrementally update representations using a continuous stream without overfitting to the recent records~\cite{zhang2017react}. When a new set of records arrives, sampling-based online learning approaches in~\cite{zhang2017react,silva2019fullustar} sample a few historical records to augment the recent records and the representations are updated using the augmented corpus to alleviate overfitting to recent records. However, sampling-based approaches need to retain historical records. To address this limitation, a constraint-based approach is proposed in~\cite{zhang2017react}, which imposes constraints on embeddings to preserve their previous embeddings. However, sampling-based approaches are superior to constraint-based approaches. In this work, we propose a novel adaptive optimization technique to accommodate online learning that mostly outperforms sampling-based approaches without storing any historical records.

\begin{figure}[t]
    \centering
    \includegraphics[width=0.9\linewidth]{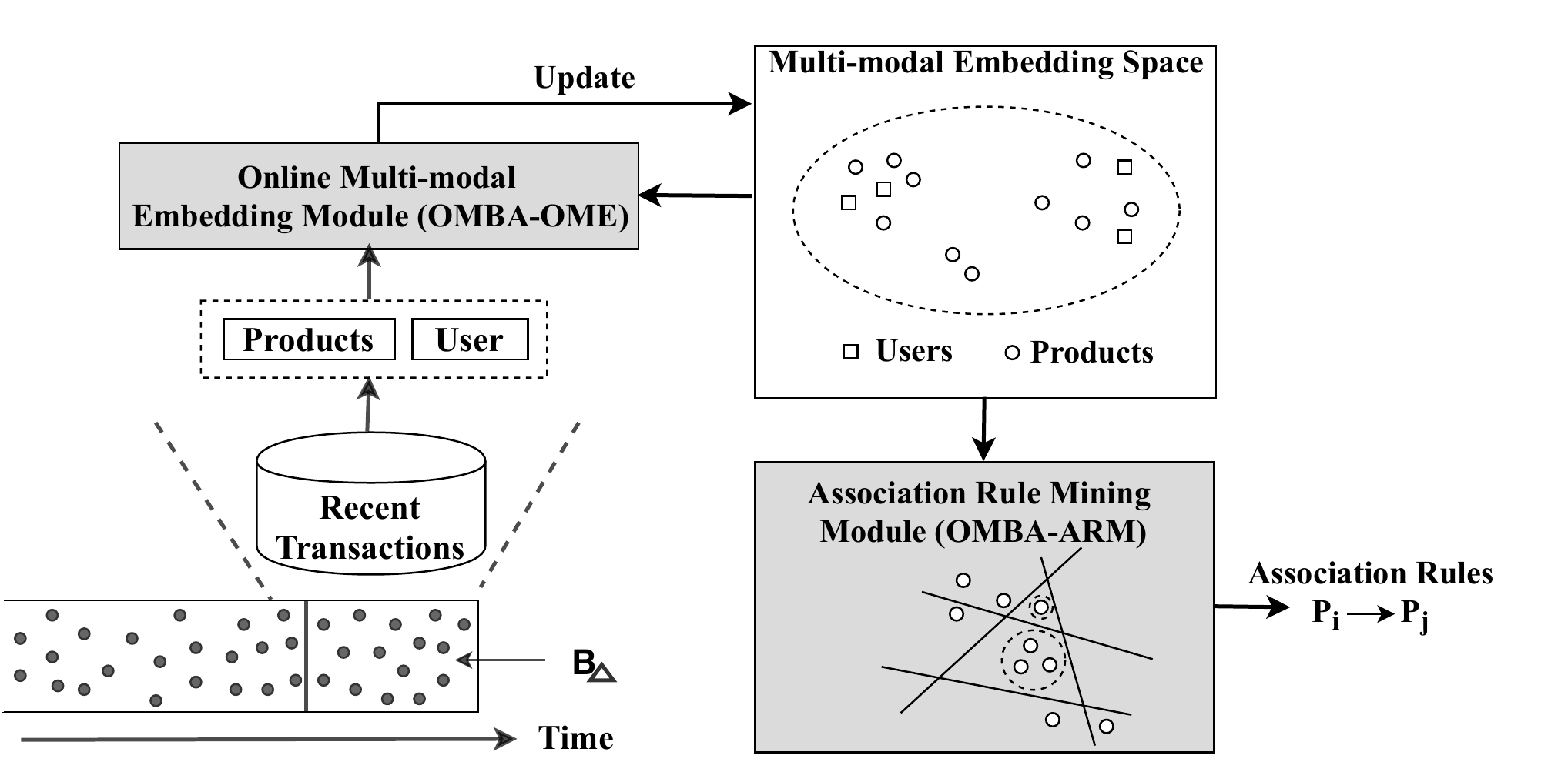}
    \caption{OMBA consists of: (1) OMBA-OME to learn online product embeddings; and (2) OMBA-ARM to generate association rules using the embeddings}
    \label{fig:model}
\end{figure}

\section{Problem Statement}\label{sec:problem}
Let $B=\{b_1, b_2, ...., b_N, ...\}$ be a continuous stream of shopping transactions (i.e., baskets) at a retailer that arrive in chronological order. Each basket $b \in B$ is a tuple $<t^b, u^b, P^b>$, where: (1) $t^b$ is the timestamp of $b$; (2) $u^b$ is the user id of $b$; and (3) $P^b$ is the set of products in $b$.

The problem is to learn the embeddings $V_P$ for $P = \bigcup^{\forall b} P^b$ (i.e., the set of unique products in $B$) such that the embedding $v_p$ of a product $p\in P$:
\begin{enumerate}[topsep=2pt]
    \item is a $d$-dimensional vector ($d << |P|$), where $|P|$ is the number of different products in $B$;
    \item preserves the associations (i.e., co-occurrences) among products; 
    \item is continuously updated as new transactions ($B_{\Delta}$) arrive to incorporate the latest information.
\end{enumerate}

\textbf{Market Basket Analysis (MBA)} uncovers the associations among products in the form of \textit{association rule}: $P_i \Rightarrow p_j$, ``Consumers who buy the products in $P_i$ are likely to buy the product $p_j$", where $P_i \subset P$ and $p_j \in P\char`\\ P_i$. The embedding learning module in \textsc{OMBA} is designed such that $p_j$ and the products in $P_i$ are mapped close together in the embedding space.

To quantitatively evaluate the embeddings that are learned for MBA, we focus on the \textbf{\textit{intra-basket item retrieval task}}, which is to retrieve the true product in a basket given the other attributes (i.e., user and other products) of the same basket. The embeddings should reflect the accurate associations between products to give better performance for this task.

\section{\textsc{OMBA}}\label{sec:model}
\subsection{Overview of \textsc{OMBA}}
OMBA consists of two modules as depicted in Figure~\ref{fig:model}: (1) Online Multi-Modal Embedding (OMBA-OME) module, which learns the representations for products; and (2) Association Rule Mining (OMBA-ARM) module, which generates the association rules using the representations from OMBA-OME.

To learn representations for products, OMBA-OME jointly embeds all the products and users into the same latent space. We conduct an empirical analysis using three datasets, which verifies that there is a significant similarity of the shopping baskets of a user (due to space limitations, more details of the empirical analysis are given in~\cite{anonymous2020omba-s}). Thus, mapping users along with the products into a single embedding space allows us to exploit the user-specific patterns in shopping baskets to generalize the product representations. 

The learning process of OMBA-OME proceeds in an online fashion, in which the embeddings are updated incrementally for the arrival of each of the new records $B_\Delta$. Such an approach avoids the unnecessary cost of learning representations from scratch for each new arrival. However, this approach could lead to overfitting to recent information while abruptly forgetting (i.e., catastrophic forgetting) the previously learned information. To address this issue, we propose an online learning method in Section~\ref{subsec:OMBA-OME}, which incorporates recent records effectively while alleviating the catastrophic forgetting without storing any historical records. Subsequently, our OMBA-ARM module adopts a clustering approach on the products' embedding space to extract association rules among the products.

\subsection{Online Multi-Modal Embedding (OMBA-OME)}\label{subsec:OMBA-OME}
\textsc{OMBA-OME} learns the embeddings for units such that the units of a given basket $b$ can be recovered by looking at $b$'s other units. Formally, we model the likelihood for the task of recovering unit $z\in b$ given the other units $b_{-z}$ of $b$ as:
\begin{equation}\label{eq:1}
    p(z|b_{-z}) = \exp (s(z,b_{-z}))/ \sum_{z'\in X} \exp (s(z', b_{-z}))
\end{equation}
$X$ is the set of units (could be user or product) of type $z$, and $s(z, b_{-z})$ is the similarity score between $z$ and $b_{-z}$. We define $s(z, b_{-z})$ as $v_z^{\top}h_z$ where, 

\begin{equation}
         h_z = 
\left\{\begin{matrix}
  (v_{u^b} + v_{\hat{P^b}})/2 & \quad \text{if $z$ is a product}\\ 
  v_{\hat{P^b}}  & \quad \text{if $z$ is a user}\\
\end{matrix}\right.
\label{eq:2}
\end{equation}

\textbf{Value-Based Weighting.} The representation for the context products  ($v_{\hat{P^b}}$) in a basket is computed based on a novel weighting scheme. This weighting scheme emphasizes learning better representations for higher value items from their rare occurrences by assigning higher weight for them considering their selling price. Formally, the term $v_{\hat{P^b}}$ in Equation~\ref{eq:2} is computed as:
\begin{equation}
    v_{\hat{P^b}} = \frac{\sum_{x \in P_{b_{-z}}} g(x) v_x}{\sum_{x \in P_{b_{-z}}} g(x)}
\end{equation}
where $g(x)$ function returns a weight for product $x$ based on its selling price, $SV(x)$. The function $g$ is computed as follows:
\begin{itemize}
\item Assuming the number of appearances of $x$ follows a power-law distribution with respect to its selling price $SV(x)$ (see Figure~\ref{fig:a}), a power-law formula (i.e., $y=cx^{-k}$) is fitted to the curve in Figure~\ref{fig:a}, which returns the probability of product $x$ appearing in a basket $b$, $p(x \in b)$, as $1.3*SV(x)^{-2.3}$ (the derivation of this formula is presented in~\cite{anonymous2020omba-s});
\item  Then, the function $g$ is computed as:
\begin{align}
    g(x) = \frac{1}{p(x\in b)}=\frac{1}{1.3*min(SV(x),10)^{-2.3}}  
\end{align}
The function $g$ is clipped for the products with selling price $>10$ ($10$ is selected as the point at which the curve in Figure~\ref{fig:a} reaches a plateau) to avoid the issue of exploding gradient.
\end{itemize}

\textbf{Adaptive Optimization. }Then, the final loss function is the negative log likelihood of recovering all the attributes of the recent transactions $B_{\Delta}$:
\begin{equation}\label{eq:pre_loss}
    O_{B_\Delta} = - \sum_{b \in B_{\Delta}} \sum_{z \in b} p(z|b_{-z})
\end{equation}

The objective function above is approximated using negative sampling (proposed in~\cite{mikolov2013distributed}) for efficient optimization. Then for a selected record $b$ and unit $z\in b$, the loss function based on the reconstruction error is:
\begin{equation}\label{eq:loss}
    L = - \log (\sigma (s(z, b_{-z}))) - \sum_{n\in N_{z}} \log (\sigma (-s(n, b_{-z})))
\end{equation}
where $N_z$ is the set of randomly selected negative units that have the type of $z$ and $\sigma(.)$ is sigmoid function.

We adopt a novel optimization strategy to optimize the loss function, which is designed to alleviate overfitting to the recent records and the frequently appearing products in the transactions.

For each basket $b$, we compute the intra-agreement $\Psi_b$ of $b$'s attributes as:
 \begin{equation}
     \Psi_b = \frac{\sum_{z_i, z_j \in b, z_i\neq z_j} \sigma(v_{z_i}^{\top}v_{z_j})}{\sum_{z_i, z_j \in b, z_i\neq z_j} 1 }
 \end{equation}
 
 Then the adaptive learning rate of $b$ is calculated as,
 \begin{equation}
     lr_b = \exp (-\tau \Psi_b)* \eta
 \end{equation}
 where $\eta$ denotes the standard learning rate and $\tau$ controls the importance given to $\Psi_b$. If the representations have already overfitted to $b$, then $\Psi_b$ takes a higher value. Consequently, a low learning rate is assigned to $b$ to avoid overfitting. In addition, the learning rate for each unit $z$ in $b$ is further weighted using the approach proposed in AdaGrad~\cite{Duchi2010} to alleviate the overfitting to frequent items. Then, the final update of the unit $z$ at the $t^{th}$ timestep is:
 \begin{equation}
     z_{t+1} = z_t - \frac{lr_b}{\sqrt{\sum_{i=0}^{t-1} {(\frac{\partial L}{\partial z})}_i^2 + \epsilon}} {(\frac{\partial L}{\partial z})}_t
 \end{equation}

\subsection{Association Rule Mining (OMBA-ARM)}\label{subsec:OMBA-ARM}
The proposed OMBA-OME module learns representations for products such that the frequently co-occurring products (i.e., products with strong associations) are closely mapped in the embedding space. Accordingly, the proposed OMBA-ARM module clusters in the products' embedding space to detect association rules between products. The product embeddings from the OMBA-OME module are in a higher-dimensional embedding space. Thus, Euclidean-distance based clustering algorithms (e.g., K-Means) suffers from the curse-of-dimensionality issue. Hence, we adopt a Locality-Sensitive Hashing (LSH) algorithm based on random projection~\cite{andoni2006near}, which assigns similar hash values to the products that are close in the embedding space. Subsequently, the products with the most similar hash values are returned as strong associations. Our algorithm is formally elaborated as the following five steps:

\begin{enumerate}
    \item Create $|F|$ different hash functions such as $F_i(v_p) = sgn(f_i\cdot v_p^{\top})$, where $i\in\{0,1, \ldots,|F|-1\}$ and $f_i$ is a $d$-dimensional vector from a Gaussian distribution $N\sim(0,1)$, and $sgn(.)$ is the sign function. According to the Johnson-Lindenstrauss lemma~\cite{johnson1984extensions}, such hash functions approximately preserve the distances between products in the original embedding space.
    \item Construct an $|F|$-dimensional hash value for $v_p$ (i.e., a product embedding) as $F_0(v_p) \oplus F_1(v_p) \oplus ... \oplus F_{|F|-1}(v_p)$, where $\oplus$ defines the concatenation operation. Compute the hash value for all the products.
    \item Group the products with similar hash values to construct a hash table. 
    \item Repeat steps (1), (2), and (3) $|H|$ times to construct $|H|$ hash tables to mitigate the contribution of bad random vectors.
    \item Return the sets of products with the highest collisions in hash tables as the products with strong associations.
\end{enumerate}
$|H|$ and $|F|$ denote the number of hash tables and number of hash functions in each table respectively.

\textbf{Finding optimal values for $|H|$ and $|F|$. }In this section, the optimal values for $|F|$ and $|H|$ are found such that they guarantee that the rules generated from \textsc{OMBA-ARM} have higher \textit{Lift}. Initially, we form two closed-form solutions to model the likelihood to have a strong association between product $x$ and product $y$ based on: (1) \textsc{OMBA-ARM}; and (2) \textsc{Lift}.

\textbf{(1) Based on \textsc{OMBA-ARM}}, products $x$ and $y$ should collide at least in a single hash table to have a strong association. Thus, the likelihood to have a strong association between product $x$ and product $y$ is\footnote{\label{note1}Detailed derivations of Eq.~\ref{eq:omba_arm} and Eq.~\ref{eq:lift} are presented in~\cite{anonymous2020omba-s}}:

\begin{align}
        p(x \Rightarrow y)_{omba} &= 1-(1 - p(sgn(v_x\cdot f) = sgn(v_y\cdot f))^{|F|})^{|H|}\notag\\
        &= 1-(1 - (1- \frac{arccos(v_x\cdot v_y)}{\pi})^{|F|})^{|H|}
    \label{eq:omba_arm}
\end{align}
where $sgn(x) = \{1\;if\; x \ge 1; -1\;otherwise\}$. $p(sgn(v_x\cdot f) = sgn(v_y\cdot f))$ is the likelihood to have a similar sign for products $x$ and $y$ with respect to a random vector. $v_x$ and $v_y$ are the normalized embeddings of product $x$ and $y$ respectively.

\textbf{(2) Based on \textsc{Lift}}, the likelihood to have a strong association between products $x$ and $y$ can be computed as\footnotemark[\getrefnumber{note1}]:

\begin{align}
    p(x \Rightarrow y)_{lift} &= \frac{Lift(y,x)_{train}}{Lift(y,x)_{train} + |N_z|*Lift(y,x)_{noise}}
    = \sigma(A*v_x\cdot v_y)
    \label{eq:lift}
\end{align}
where $Lift(y,x)_{train}$ and $Lift(y,x)_{noise}$ are the \textit{Lift} scores of $x$ and $y$ calculated using the empirical distribution of the dataset and the noise distribution (to sample negative samples) respectively. $|N_z|$ is the number of negative samples for each genuine sample in Equation~\ref{eq:loss}, and $A$ is a scalar constant.

Then, the integer solutions for parameters $|F|=4$ and $|H|=11$, and real solutions for $A=4.3$ are found such that $p(x \Rightarrow y)_{omba} = p(x \Rightarrow y)_{lift}$. Such a selection of hyper-parameters theoretically guarantees that the rules produced from \textsc{OMBA-ARM} are higher in \textit{Lift}, which is a statistically well-defined measure for strong associations. 

The advantages of our OMBA-ARM module can be listed as follows: (1) it is simple and efficient, which simplifies the expensive comparison of all the products in the original embedding space ($O(dN^2)$ complexity) to $O(d|F||H|N)$, where $N\gg|F||H|$; (2) it approximately preserves the distances in the original embedding space, as the solutions for $|F|(=4)$ and $|H|(=11)$ satisfy the  Johnson–Lindenstrauss lemma~\cite{johnson1984extensions} $(|F||H|\gg log(N))$; (3) it outperforms underlying Euclidean norm-based clustering algorithms (e.g., K-means) for clustering data in higher dimensional space, primarily due to the curse of dimensionality issue
~\cite{aggarwal2001surprising}; and (4) theoretically guarantees that the rules have higher \textit{Lift}. 

\section{Experimental Methodology}\label{sec:experiments}
\textbf{Datasets.} We conduct all our experiments using three publicly available real-world datasets:

\begin{itemize}[topsep=2pt]
\item Complete Journey Dataset
 (CJ) contains household-level transactions at a retailer by 2,500 frequent shoppers over two years.
\item Ta-Feng Dataset
 (TF) includes shopping transactions of the Ta-Feng supermarket from November 2000 to February 2001.
\item InstaCart Dataset
 (IC) contains the shopping transactions of Instacart, an online grocery shopping center, in 2017.
\end{itemize}
The descriptive statistics of the datasets are shown in Table~\ref{tab:dataset_statistics}. As can be seen, the datasets have significantly different statistics (e.g., TF has a shorter collection period, and IC has a larger user base), which helps to evaluate the performance in different environment settings. The time-space is divided into $1$-day time windows, meaning that new records are received by the model once per day.

\begin{table}[t]
\scriptsize
\centering
\caption{Descriptive statistics of the datasets}\label{tab:dataset_statistics} 
\begin{tabular}{l|r|r|r|r}
\hline
Datasets & \# Users &   \# Items& \# Transactions & \# Baskets \\
\hline
\hline
Complete Journey (CJ) & 2,500& 92,339  & 2,595,733& 276,483\\
\hline
Ta-Feng (TF) & 9,238 & 7,973 & 464,118 & 77,202\\
\hline
Instacart (IC) &206,209 & 49688 &33,819,306 & 3,421,083\\
\hline
\end{tabular}
\end{table}

\noindent \textbf{Baselines. }We compare \textsc{OMBA} with the following methods:
\begin{itemize}[topsep=2pt]
    \item \textsc{Pop} recommends the most popular products in the training dataset. This is shown to be a strong baseline in most of the domains, despite its simplicity. 
    \item \textsc{Sup} recommends the product with the highest \textit{Support} for a given context.
    \item \textsc{Lift} recommends the product that has the highest \textit{Lift} for a given context.
    \item \textsc{NMF}~\cite{lee2001algorithms} performs Non-Negative Matrix Factorization on the user-product co-occurances matrix to learn representations for users and products.
    \item \textsc{Item2Vec}~\cite{barkan2016item2vec} adopts word2vec for learning the product representations by considering a basket as a sentence and a product in the basket as a word in the sentence.
    \item \textsc{Prod2Vec}~\cite{grbovic2015commerce} is the \textsc{Bagged-Prod2Vec} version, which aggregates all the baskets related to a single user as a product sequence and apply word2vec to learn representations for products.
    \item \textsc{Triple2Vec}~\cite{wan2018representing} joinlty learns representations for products and users such that they preserve the triplets in the form of $<user, item_1, item_2>$, which are generated using shopping baskets. 
\end{itemize}
Also, we compare with a few variants of \textsc{OMBA}.
\begin{itemize}[topsep=2pt]
    \item \textsc{OMBA-No-User} does not consider users. The comparison with \textsc{OMBA-No-User} highlights the importance of users.
    \item \textsc{OMBA-Cons} adopts SGD optimization with the constraint-based online learning approach proposed in~\cite{zhang2017react}.
    \item \textsc{OMBA-Decay} and \textsc{OMBA-Info} adopt SGD optimization with the sampling-based online learning methods proposed in~\cite{zhang2017react} and~\cite{silva2019fullustar} respectively.
\end{itemize}

\noindent\textbf{Parameter Settings. }All the representation learning based techniques share three common parameters (default values are given in brackets): (1) the latent embedding dimension $d\text{ }(300)$, (2) the SGD learning rate $\eta\text{ }(0.05)$, (3) the negative samples $|N_z|\text{ }(3)$ as appeared in Equation~\ref{eq:loss}, and (4) the number of epochs $N\text{ }(50)$. We set $\tau=0.1$ after performing grid search using the CJ dataset (see~\cite{anonymous2020omba-s} for a detailed study of parameter sensitivity). For the specific parameters of the baselines, we use the default parameters mentioned in their original papers.

\noindent\textbf{Evaluation Metric. }Following the previous work~\cite{zhang2017react}, we adopt the following procedure to evaluate the performance for the \textit{intra-basket item retrieval task}. For each transaction in the test set, we select one product as the target prediction and the rest of the products and the user of the transaction as the context. We mix the ground truth target product with a set of $M$ negative samples (i.e., products) to generate a candidate pool to rank. $M$ is set to 10 for all the experiments. Then the size-$(M+1)$ candidate pool is sorted to get the rank of the ground truth. The average similarity of each candidate product to the context of the corresponding test instance is used to produce the ranking of the candidate pool. Cosine similarity is used as the similarity measure of \textsc{OMBA}, \textsc{Triple2Vec}, \textsc{Prod2Vec}, \textsc{Item2Vec} and \textsc{NMF}. \textsc{POP}, \textsc{Sup}, and \textsc{Lift} use popularity, support, and lift scores as similarity measures respectively. 

If the model is well trained, then higher ranked units are most likely to be the ground truth. Hence, we use three different evaluation metrics to analyze the ranking performance: (1) Mean Reciprocal Rank (MRR) $ =\frac{\sum_{q = 1}^Q 1/rank_i}{|Q|}$; (2) Recall@k (R@k) $ =\frac{\sum_{q = 1}^Q min(1, \lfloor k/rank_i \rfloor)}{|Q|}$; and (3) Discounted Cumulative Gain (DCG) $ =\frac{\sum_{q = 1}^Q 1/log_2(rank_i+1)}{|Q|}$, where $Q$ is the set of test queries and $rank_i$ refers the rank of the ground truth label for the $i$-th query. $\lfloor.\rfloor$ is the floor operation. A good ranking performance should yield higher values for all three metrics. We randomly select 20 one-day query windows from the second half of the period for each dataset, and all the transactions in the randomly selected time windows are used as test instances. For each query window, we only use the transactions that arrive before the query window to train different models. Only \textsc{OMBA} and its variants are trained in an online fashion and all the other baselines are trained in a batch fashion for 20 repetitions.

\section{Results}\label{sec:results}

\begin{sidewaystable}
\scriptsize
    \caption{The comparison of different methods for \textit{intra-basket item retrieval}. Each model is evaluated 5 times with different random seeds and the mean value for each model is presented. Recall values for different $k$ values are presented in~\cite{anonymous2020omba-s} due to space limitations}
    \centering
    \begin{tabular}{|l|c|c|c|c|c|c|c|c|c|c|c|c|c|c|c|c|c|}
    \cline{2-11}
    \multicolumn{1}{c|}{}& \multicolumn{9}{c|}{Results for \textit{intra-basket item retrieval}} & \multirow{3}{*}{Memory Complexity}\\
    \cline{1-10}
    Dataset &  \multicolumn{3}{c|}{CJ} & \multicolumn{3}{c|}{IC} & \multicolumn{3}{c|}{TF} &\\
    \cline{1-10}
    Metric & MRR& R@1 &DCG& MRR& R@1 &DCG& MRR& R@1 &DCG&\\
    \hline
    \textsc{Pop} &0.2651&0.095&0.4295&0.2637&0.07841&0.4272&0.2603&0.0806&0.4247& $O(P + |B_{max}|)$\\
    \textsc{Sup} &0.3308&0.1441&0.4839&0.3009&0.1061&0.4634&0.3475&0.1646&0.4972& $O(P^2 + |B_{max}|)$\\
    \textsc{Lift} &0.5441&0.3776&0.6477&0.4817&0.2655&0.6170&0.4610&0.2981&0.5868& $O(P(P+1) + |B_{max}|)$\\
    \hline \hline
    \textsc{NMF} &0.1670&0.0000&0.3565&0.5921&0.3962&0.6922&0.4261&0.2448&0.5591& $O(P(U+P) + |B_{max}|)$\\
    \textsc{Item2Vec} &0.4087&0.2146&0.5457&0.5159&0.2929&0.6137&0.3697&0.1782&0.5149& $O(k(2*P) + |B|)$\\
    \textsc{Prod2Vec} &0.4234&0.2275&0.5575&0.5223&0.3222&0.6363&0.3764&0.1854&0.5201& $O(k(2*P) + |B|)$\\
    \textsc{Triple2Vec} &0.5133&0.3392&0.6269&0.6169&0.4348&0.7095&0.478&0.3040&0.5990& $O(k(U+2*P) + |B|)$\\
    \hline \hline
    \textsc{OMBA-No-User} &0.4889&0.3091&0.6004&0.5310&0.3384&0.6405&0.3873&0.2013&0.5298& $O(kP + |B_{max}|)$\\
    \textsc{OMBA-Cons} &0.4610&0.2742&0.5843&0.5942&0.3393&0.6998&0.3996&0.2031&0.5324& $O(k(U+P) + |B_{max}|)$\\
    \textsc{OMBA-Decay} &0.5984&0.4221&0.6948&0.7117&0.5442&0.7860&0.402&0.2186&0.5387& $O(k(U+P) + |B_{max}|/(1-e^{-\tau}))$\\
    \textsc{OMBA-Info} &0.5991&0.4275&0.6937&\textbf{0.7482}&0.5852&\textbf{0.8027}&0.4046&0.2205&0.5421& $O(k(U+P) + |B_{max}|/(1-e^{-\tau}))$\\
    \hline \hline
    \textsc{OMBA} &\textbf{0.6013}&\textbf{0.4325}&\textbf{0.6961}&0.7478&\textbf{0.5859}&0.8025&\textbf{0.5166}&\textbf{0.3466}&\textbf{0.6293}& $O(k(U+P) + |B_{max}|)$\\
    \hline
    \end{tabular}
    \label{tab:results1}
    \vspace{2\baselineskip}
    \caption{(a) 5 nearest neighbours in the embedding space of \textsc{OMBA} and \textsc{Triple2Vec} for a set of target products; (b) MRR for \textit{intra-basket item retrieval} using the test queries, such that the ground truth's price $> x$ (in dollars). All the results are calculated using CJ}\label{fig:results_1_2}
    \subfloat[]{%
    \centering
    \begin{tabular}{|c|c|c|}
    \hline
    \begin{tabular}[c]{@{}c@{}}Target\\ Product\end{tabular}&                  \begin{tabular}[c]{@{}c@{}}5 nearest\\ products by \textsc{OMBA} \end{tabular}                                                                                                                                 & \begin{tabular}[c]{@{}c@{}}5 most nearest\\ products by \textsc{Triple2Vec}\end{tabular} \\ \hline \hline
         \begin{tabular}[c]{@{}c@{}}Layer \\Cakes\end{tabular}   
    & \begin{tabular}[c]{@{}c@{}}Cake Candles, Cake Sheets,\\ Cake Decorations, Cake Novelties,\\ Birthday/Celebration Layer \end{tabular}
    & \begin{tabular}[c]{@{}c@{}}Cheesecake, Fruit/Nut Pies,\\ Cake Candles, Flags,\\ Salad Ingredients\end{tabular}\\ \hline
        \begin{tabular}[c]{@{}c@{}}Frozen\\ Bread\end{tabular}                       
    & \begin{tabular}[c]{@{}c@{}}Sauces, Eggs, Peanut\\ Butter, Non Carbohydrate\\ Juice, Pasta/Ramen \end{tabular}
    & \begin{tabular}[c]{@{}c@{}}Frozen Breakfast, Breakfast Bars,\\ Popcorn, Gluten Free Bread,\\ Cookies/Sweet Goods\end{tabular}                                                      \\ \hline
    \begin{tabular}[c]{@{}c@{}}Turkey\end{tabular}                       
    & \begin{tabular}[c]{@{}c@{}}Beef, Meat Sauce,\\ Oil/Vinegar, Cheese, Ham\end{tabular}
    & \begin{tabular}[c]{@{}c@{}}Ham, Beef, Cheese,\\ Chicken, Lunch Meat\end{tabular}                                                      \\ \hline
    \begin{tabular}[c]{@{}c@{}}Authentic\\ Indian\\ Foods\end{tabular}                       
    & \begin{tabular}[c]{@{}c@{}}Grain Mixes, Organic Pepper,\\ Other Asian Foods, Tofu,\\ South American Wines\end{tabular}
    & \begin{tabular}[c]{@{}c@{}}Other Asian Foods, German \\Foods, Premium Mums, Herbs\\ \& Fresh Others, Bulb Sets\end{tabular}\\ \hline
                        
    \end{tabular}%
\label{fig:similar_prod}
}\hspace{1em}
\subfloat[]{%
\scriptsize
\centering
\begin{tikzpicture}
\begin{axis}[
    width=7.5cm,
    height=5cm,
    ylabel={MRR},
    xlabel={$x$},
    ymin=0.555, ymax=0.61,
    xtick={0, 2, 4, 6, 8, 10},
    legend pos=south east,
    ymajorgrids=true,
    grid style=dashed,
]
\addplot[
    color=black,
    mark=*
    ]
    coordinates {
    (0,0.6013)(2,0.5973)(4,0.5983)(6,0.5987)(8,0.5994)(10,0.5998)
    };
\addplot[
    color=black,
    mark=*, 
    dashdotted,
    ]
    coordinates {
    (0,0.5909)(2,0.5887)(4,0.5800)(6,0.5795)(8,0.5765)(10,0.5743)
    };
    \legend{\textsc{OMBA}, \textsc{OMBA-W/O Value Based Weight.}, \textsc{Triple2Vec}}
\end{axis}
\end{tikzpicture}
\label{fig:value_weighting}
}%

\end{sidewaystable}

\textbf{Intra-basket Item Retrieval Task. }Table~\ref{tab:results1} shows the results collected for the \textit{intra-basket item retrieval} task. \textsc{OMBA} and its variants show significantly better results than the baselines, outperforming the best baselines on each dataset by as much as 10.51\% for CJ, 21.28\% for IC, and 8.07\% for TF in MRR. 

\textbf{(1) Capture semantics of products. }\textsc{Lift} is shown to be a strong baseline. However, it performs poorly for IC, which is much sparser with a large userbase compared to the other datasets. This observation clearly shows the importance of using representation learning based techniques to overcome data sparsity by capturing the semantics of products. We further analyzed the wrongly predicted test instances of \textsc{Lift}, and observed that most of these instances are products with significant seasonal variations such as ``Rainier Cherries'' and ``Valentine Gifts and Decorations'' (as shown in Figure~\ref{fig:c}). This shows the importance of capturing temporal changes of the products' associations.

Out of the representation learning-based techniques, \textsc{Triple2Vec} is the strongest baseline, despite being substantially outperformed by \textsc{OMBA}. As elaborated in Section~\ref{sec:related}, \textsc{Triple2Vec} mainly preserves semantic similarity between products. This could be the main reason for the performance difference between \textsc{Triple2Vec} and \textsc{OMBA}. To validate that, Table~\ref{fig:similar_prod} lists the nearest neighbours for a set of products in the embedding spaces from \textsc{OMBA} and \textsc{Triple2Vec}. Most of the nearest neighbours in \textsc{Triple2Vec} are semantically similar to the target products. For example, `Turkey' has substitute products like `Beef' and `Chicken' as its nearest neighbours. In contrast, the embedding space of \textsc{OMBA} mainly preserve complementarity. Thus, multiple related products to fulfill a specific need are closely mapped. For example, `Layer Cake' has neighbours related to a celebration (e.g., Cake Decorations and Cake Candles).

\textbf{(2) Value-based weighting. }To validate the importance of the proposed value-based weighting scheme in \textsc{OMBA}, Table~\ref{fig:value_weighting} shows the deviation of the performance for \textit{intra-basket item retrieval} with respect to the selling price of the ground truth products. With the proposed value-based weighting scheme, \textsc{OMBA} accurately retrieves the ground truth products that have higher selling prices. Thus, we can conclude that the proposed value-based weighting scheme is important to learn accurate representations for rarely occurring products.

\textbf{(3) Online learning. }Comparing the variants of \textsc{OMBA}, \textsc{OMBA} substantially outperforms \textsc{OMBA-No-User}, showing that users are important to model products' associations. \textsc{OMBA}'s results are comparable (except 27.68\% performance boost for TF) with sampling-based online learning variants of \textsc{OMBA} (i.e., \textsc{OMBA-Decay} and \textsc{OMBA-Info}), which store historical records to avoid overfitting to recent records. Hence, the proposed adaptive optimization-based online learning technique in \textsc{OMBA} achieves the performance of the state-of-the-art online learning methods (i.e., \textsc{OMBA-Decay} and \textsc{OMBA-Info}) in a memory-efficient manner without storing any historical records. 

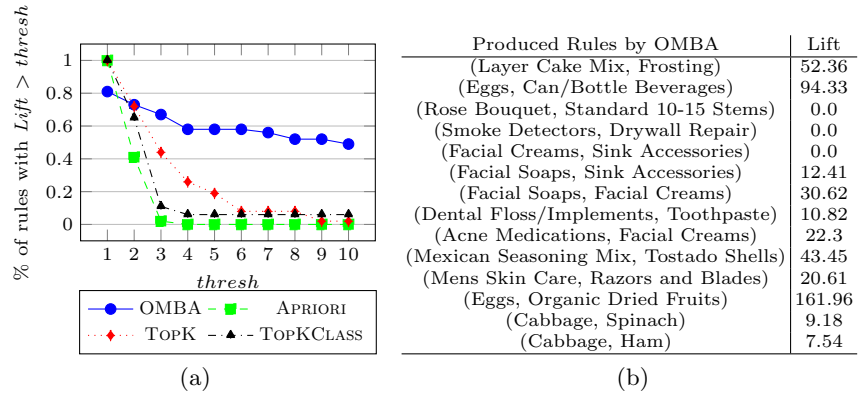
\begin{figure*}[t]
\scriptsize
\centering
\subfloat[]{%
\begin{tikzpicture}

\begin{axis}[
    width=5.5cm,
    height=4.2cm,
    xlabel={$thresh$},
    ylabel={\% of rules with \textit{Lift} $>thresh$},
    xmin=0, xmax=11,
    ymin=-0.1, ymax=1.1,
    xtick={1, 2, 3, 4, 5, 6, 7, 8, 9, 10},
    ytick={0.0, 0.2, 0.4, 0.6, 0.8, 1.0},
    legend style={at={(0.5,-0.25)},anchor=north,legend columns=2},
    ymajorgrids=true,
]
\addplot[
    color=blue,
    mark=*,
    ]
    coordinates {
    (1,0.81)(2,0.73)(3,0.67)(4,0.58)(5,0.58)(6,0.58)(7,0.56)(8,0.52)(9,0.52)(10,0.49)
    };
\addplot[
    color=green,
    mark=square*,
    dashed,
    ]
    coordinates {
    (1,1)(2,0.41)(3,0.02)(4,0)(5,0)(6,0)(7,0)(8,0)(9,0)(10,0)
    };
\addplot[
    color=red,
    mark=diamond*, 
    dotted,
    ]
    coordinates {
    (1,1)(2,0.72)(3,0.44)(4,0.26)(5,0.19)(6,0.08)(7,0.08)(8,0.08)(9,0.02)(10,0.02)
    };
\addplot[
    color=black,
    mark=triangle*, 
    dashdotted,
    ]
    coordinates {
    (1,1)(2,0.65)(3,0.11)(4,0.06)(5,0.06)(6,0.06)(7,0.06)(8,0.06)(9,0.06)(10,0.06)
    };
    \legend{\textsc{OMBA}, \textsc{Apriori}, \textsc{TopK}, \textsc{TopKClass}}
\end{axis}
\end{tikzpicture}
\label{fig2:a}
}\hspace{0.5em}
\subfloat[]{%
\begin{tabular}[b]{c|c}
        Produced Rules by OMBA & Lift\\
        \hline
        (Layer Cake Mix, Frosting) & 52.36\\
        (Eggs, Can/Bottle Beverages) & 94.33\\
         (Rose Bouquet, Standard 10-15 Stems) & 0.0 \\
         (Smoke Detectors, Drywall Repair) & 0.0 \\
         (Facial Creams, Sink Accessories) & 0.0 \\
         (Facial Soaps, Sink Accessories) & 12.41\\
         (Facial Soaps, Facial Creams) & 30.62\\
         (Dental Floss/Implements, Toothpaste) & 10.82 \\
         (Acne Medications, Facial Creams) & 22.3\\
         
         (Mexican Seasoning Mix, Tostado Shells) & 43.45\\
         (Mens Skin Care, Razors and Blades) & 20.61\\
         (Eggs, Organic Dried Fruits) & 161.96\\
         (Cabbage, Spinach) & 9.18\\
         (Cabbage, Ham) & 7.54\\
         \hline
\end{tabular}
\label{fig2:b}
}%
\caption{(a) The percentage of top 100 rules that have \textit{Lift} scores $>thresh$; and (b) A few examples for the top 100 rules produced by OMBA} \label{fig:results2}
\end{figure*}

\textbf{Association Rule Mining. }To compare the association rules generated from OMBA, we generate the association rules for the CJ dataset using:
(1) Apriori algorithm~\cite{agrawal1994fast} with \textit{minimum Support=0.0001}; (2) TopK Algorithm~\cite{fournier2012mining} with \textit{k=100}; and (3) TopKClass Algorithm\footnote{\url{https://bit.ly/spmf_TopKClassAssociationRules}} with \textit{k=100} and \textit{consequent list= \{products appear in the rules from \textsc{OMBA}}\}. Figure~\ref{fig2:a} shows \textit{Lift} scores of the top 100 rules generated from each approach. 

\textbf{(1) Emphasize rarely occurring associations.} As can be seen, at most 20\% of the rules generated from the baselines have \textit{Lift} scores greater than 5, while around 60\% of the rules of \textsc{OMBA} have scores greater than 5. This shows that \textsc{OMBA} generates association rules that are strong and rarely occurring in the shopping baskets (typically have high \textit{Lift} scores due to low \textit{Support} values), which are not emphasized in conventional MBA techniques.

\begin{table}[t]
    \centering
    \scriptsize
    \caption{Association Rules generated by \textsc{OMBA-ARM} on different days}
    \begin{tabular}{|c|c|c|}
    \cline{2-3}
    \multicolumn{1}{c|}{}&                  \multicolumn{2}{c|}{\begin{tabular}[c]{@{}c@{}}Mostly Associated Products by \textsc{OMBA-ARM} \end{tabular}}   \\
    \hline
    \begin{tabular}[c]{@{}c@{}}Target Product\end{tabular}& At Month 11 & At Month 17\\
    \hline
    \begin{tabular}[c]{@{}c@{}}Valentine Gifts \\and Decorations\end{tabular}& \begin{tabular}[c]{@{}c@{}}Rose Consumer Bunch, Apple Juice,\\ Ladies Casual Shoes\end{tabular} & \begin{tabular}[c]{@{}c@{}}Peaches, Dried Plums,\\ Drain Care\end{tabular}\\
    \hline
    \begin{tabular}[c]{@{}c@{}}Mangoes\end{tabular}& \begin{tabular}[c]{@{}c@{}}Blueberries, Dry Snacks, Raspberries\end{tabular} & \begin{tabular}[c]{@{}c@{}}Dry Snacks, Red Cherries, Tangelos\\\end{tabular}\\
    \hline
    \begin{tabular}[c]{@{}c@{}}Frozen Bread \end{tabular}& \begin{tabular}[c]{@{}c@{}}Ham, Fluid Milk, Eggs\end{tabular} & \begin{tabular}[c]{@{}c@{}}Peanut Butter, Eggs, Ham\end{tabular}\\
    \hline
    \end{tabular}
    \label{tab:temporal_associations}
\end{table}

\textbf{(2) Discover unseen associations.} However, Figure~\ref{fig2:a} shows that some of the rules from \textsc{OMBA} have \textit{Lift} scores less than 1 (around 20\% of the rules). Figure~\ref{fig2:b} lists some of the rules generated from \textsc{OMBA}, including a few with $Lift < 1$. The qualitative analysis of these results shows that OMBA can even discover unseen associations in the dataset (i.e., associations with \textit{Lift}$=0$) as shown in Figure~\ref{fig2:b}, by capturing the semantics of product names. The associations with \textit{Lift}$=0$ like ``(Rose Bouquet, Standard 10-15 Stems)'' and ``(Smoke Detectors, Drywall Repair)'' can be interpreted intuitively, but they would be omitted by conventional count-based approaches. The ability to capture the semantics of products enables \textsc{OMBA} to mitigate the \textit{cold-start problem} and recommend products to customers that they have never purchased. The role of the OMBA-OME module on capturing products' semantics can be further illustrated using the rule ``(Facial Creams, Sink Accessories)'', which is unseen in the CJ dataset. However, both ``Sink Accessories'' and ``Facial Creams'' frequently co-occurred with ``Facial Soaps'', with 12.41 and 30.02 \textit{Lift} scores. Hence, the representations for ``Sink Accessories'' and ``Facial Creams'' learnt by \textsc{OMBA-OME} are similar to the representation of ``Facial Soaps'' to preserve their co-occurrences. This leads to similar representations for ``Sink Accessories'' and ``Facial Creams'' and generates the rule. In contrast, conventional count-based approaches (e.g., Apriori, H-Mine, and TopK) are unable to capture these semantics of product names as they consider each product as an independent unit.

\textbf{(3) Temporal variations of associations.} Moreover, Table~\ref{tab:temporal_associations} shows that \textsc{OMBA} adopts associations rules to the temporal variations. For example, most of the associated products for `Mangoes' at month 11 are summer fruits (e.g., `Blueberries' and `Raspberries'); `Mangoes' are strongly associated with winter fruits (e.g., `Cherries' and `Tangelos') at Month 17. However, everyday products like `Frozen Bread' show consistent associations with other products. These results further signify the importance of learning online product representation. 

In summary, the advantages of \textsc{OMBA} are three-fold: (1) \textsc{OMBA} explores the whole products' space effectively to detect rarely-occurring non-trivial strong associations; (2) \textsc{OMBA} considers the semantics between products when learning associations, which alleviates the cold-start problem; and (3) \textsc{OMBA} learns products' representations in an online fashion, thus is able to capture the temporal changes of the products' associations accurately. 

\section{Conclusion}\label{sec:conclusion}
We proposed a scalable and effective method to perform Market Basket Analysis in an online fashion. Our model, \textsc{OMBA}, introduced a novel online representation learning technique to jointly learn embeddings for users and products such that they preserve their associations. Following that, \textsc{OMBA} adapted an effective clustering approach in high dimensional space to generate association rules from the embeddings. Our results show that \textsc{OMBA} manages to uncover non-trivial and temporally changing association rules at the finest product levels, as \textsc{OMBA} can capture the semantics of products in an online fashion. Nevertheless, extending \textsc{OMBA} to perform MBA for multi-level products and multi-store environments could be a promising future direction to alleviate data sparsity at the finest product levels. Moreover, it is worthwhile to explore sophisticated methods to incorporate users' purchasing behavior to improve product representations. 

%
%

%
%
%
%

\end{document}


\title{Supplementary Materials for OMBA: User-Guided Product Representations for Online Market Basket Analysis}
\titlerunning{Item Representations for Online Market Basket Analysis}
%
\author{Amila Silva \and
Ling Luo \and
Shanika Karunasekera \and 
Christopher Leckie}
\authorrunning{Amila Silva et al.}
%
\institute{School of Computing and Information Systems\\ The University of Melbourne \\Parkville, Victoria, Australia\\
\email{\{amila.silva@student., ling.luo@, karus@, caleckie@\}unimelb.edu.au}}
\maketitle              
\vspace{-4mm}
\begin{abstract}
This is the supplementary material for OMBA: User-Guided Product Representations for Online Market Basket Analysis~\cite{anonymous2020omba}. The structure of this material is as follows. In Section~\ref{sec:1}, we present detailed results of the empirical analysis conducted to verify users' repetitive purchasing behavior. Section~\ref{sec:2} provides the derivation of the relationship between products' selling price and their frequency of occurrence in shopping baskets. Section~\ref{sec:3} theoretically proves that the rules generated from OMBA-ARM have higher Lift with the correct selection of hyper-parameters. We provide detailed results for the hyper-parameter sensitivity of \textsc{OMBA} in Section~\ref{sec:4}. More results on intra-basket item retrieval tasks are presented in Section~\ref{sec:5}.
\keywords{Market Basket Analysis \and Online Learning \and Item Representations \and Transaction Data}
\end{abstract}

\section{Empirical Analysis on Users' Purchasing Behavior}\label{sec:1}

In this analysis, the following question is investigated: \textit{Do users tend to buy the same items over and over again?}. If this is true, users exhibit repeating buying patterns and the historical transactions of a user could be useful to predict the user's future transactions. We conduct the following test to explore the aforementioned research question:

\begin{enumerate}
\item For each shopping basket $b$, represent the products in $b$ by a TFIDF vector $I_p^b$ (considering transactions as documents and products as words).
\item\label{lis:step2} Randomly sample a user $u$, who have at least two transactions. Sample two transactions, $b_x$ and $b_y$, from $u$'s transaction history and calculate the similarity of $b_x$ and $b_y$ using cosine similarity between $I_p^{b_x}$ and the $I_p^{b_y}$
\item\label{lis:step3} Perform Step~\ref{lis:step2} $k$ times to construct the set of similarity scores, $S_{same\_users}$, which denotes the similarity of the transactions of the same user.
\item Similarly, perform Step~\ref{lis:step2} iteratively $k$ times with the tweet pairs from different users to construct $S_{diff\_users}$.
\item Test hypothesis $H_0: \overline{S_{same\_users}} \le \overline{S_{diff\_users}}$ statistically (using one-tailed t-test) to verify the research question ($H_0$ should be rejected to verify the question), where $\overline{S_{same\_users}}$ and $\overline{S_{diff\_users}}$ denote the mean values of $S_{same\_users}$ and $S_{diff\_users}$ respectively. 
\end{enumerate}
We experiment with $k=100,000$ using the finest product level category of Complete Journey (CJ), a publicly available real-world dataset (the statistics are shown in Table~\ref{tab:1}). For $k\ge100,000$,   $\overline{S_{same\_users}}$ and $\overline{S_{diff\_users}}$ remain stable around $0.0575$ and $0.0043$ respectively. This results in an almost zero p-value ($t_{stat} = 82.4665, p_{val} = 0.0$) for the aforementioned t-test, which verifies the users' tendency to buy same set of products repeatedly.

\begin{table}[t]
\scriptsize
\centering
\caption{Descriptive statistics of the datasets}\label{tab:1} 
\begin{tabular}{l|r|r|r|r}
\hline
Datasets & \# Users &   \# Items& \# Transactions & \# Baskets \\
\hline
\hline
Complete Journey (CJ) & 2,500& 92,339  & 2,595,733& 276,483\\
\hline
\end{tabular}
\vspace{-2mm}
\end{table}

\section{Product's Selling Price vs Frequency of Occurrence}\label{sec:2}

\begin{figure*}[b]
\scriptsize
\centering
\subfloat[]{%
\begin{tikzpicture}
\label{fig:1_a}
\begin{axis}[
    width=0.42\linewidth,
    xlabel={$product\text{ }price\text{ }(in\text{ }dollars)$},
    ylabel={$\#\text{ }of\text{ }individual\text{ }sales$},
    xtick={0, 10, 20, 30, 40, 50},
    legend pos=south east,
    ymajorgrids=true,
    grid style=dashed,
]
\addplot[
    color=blue,
    each nth point={1},
    line width= 0.5pt,
    ] table[x=t,y=v,col sep=comma] {ex4.csv};
\end{axis}
\end{tikzpicture}%
}\hspace{1em}
\subfloat[]{%
\begin{tikzpicture}
\label{fig:1_b}
\begin{axis}[
    width= 0.5\linewidth,
    height= 4.5cm,
    xlabel={$ln(SV(x))$},
    ylabel={$ln(h(x))$},
    xtick={0,1,2,3,4},
    legend pos=north east,
    ymajorgrids=true,
    grid style=dashed,
    color=black
]
\addplot[
    only marks,
    mark=*,
    mark size=1.5pt,
    color=black
    ] table[x=t,y=v,col sep=comma] {ex6.csv};
\addplot[color = black, domain=0:4] {15.5-2.3*x};
\legend{$actual\text{ }points$,$fitted\text{ }curve$}
\end{axis}
\end{tikzpicture}
}
\vspace{-1mm}
\caption{(a) Number of products' sales with respect to the products' prices in CJ dataset; (b) Fitted linear curve, $ln(h(x))=15.5 -2.3*ln(SV(x))$, for Equation~\ref{eq:linear} assuming Figure~\ref{fig:1_a} follows a power-law distribution}\label{fig:example1}
\vspace{-6mm}
\end{figure*}

In this section, we derive a relationship between the price of a product and its likelihood to appear in a transaction. 

Initially, we assume that the number of appearances, $h(x)$, of product $x$ follows a power-law distribution with respect to its selling price $SV(x)\text{ }(in \text{ }dollars)$ (see Figure~\ref{fig:1_a}). Based on this assumption, a power-law formula is fitted to the curve in Figure~\ref{fig:1_a}, which can be simplified to a linear regression problem as follows. 
\begin{align}
    h(x) &= p*SV(x)^q\notag\\
    ln(h(x)) &= ln(p) + q*ln(SV(x))
    \label{eq:linear}
\end{align}

By fitting Equation~\ref{eq:linear} using the curve in Figure~\ref{fig:1_a}, $p$ and $q$ can be found as $5.4*10^6$ and $-2.3$ respectively (see Figure~\ref{fig:1_b} for the fitted curve).
For a given shopping basket $b$, the probability to have product $x$ of price $SV(x)$ is computed as:
\begin{align}
    p(x \in b)
    &= \frac{5.4*10^6*SV(x)^{-2.3}}{\int_{0}^{\inf} 5.4*10^6*SV(x_0)^{-2.3} d(SV(x_0))}\notag\\
     &= 1.3SV(x)^{-2.3}
\end{align}

\section{Finding Optimal $|F|$ and $|H|$ in \textsc{OMBA-ARM}}\label{sec:3}

In this section, the optimal values for $|F|$ and $|H|$ (see Section 4.3 in~\cite{anonymous2020omba} to learn about these parameters) are found such that they establish a theoretical relationship between the association rules generated from \textsc{OMBA-ARM} and \textsc{Lift}. Initially, we form two closed-form solutions to model the likelihood to have a strong association between product $x$ and product $y$ based on: (1) \textsc{OMBA-ARM}; and (2) \textsc{Lift}.

\textbf{(1) Based on \textsc{OMBA-ARM}.} products $x$ and $y$ should collide at least in a single hash table to have a strong association. Thus, the likelihood to have a strong association between product $x$ and product $y$ is:

\begin{equation}
        p(x \Rightarrow y)_{omba} = 1-(1 - p(sgn(v_x\cdot f) = sgn(v_y \cdot f))^{|F|})^{|H|}
    \label{eq:omba_arm_app_1}
\end{equation}
where $sgn(x) = 1\;if\; x \ge 1; and -1\;otherwise$. $p(sgn(v_x \cdot f) = sgn(v_y \cdot f))$ (i.e., the likelihood to have a similar hash value for products $x$ and $y$ with respect to a random vector) can be computed from the following lemma. 

\begin{lemma}\label{lemma:1_app}
For a given pair of normalized vectors $v_x$ and $v_y$, the likelihood to produce similar sign values from their signed projections with respect to a random vector $f$ is:
\begin{equation*}
    p(sgn(v_x\cdot f) = sgn(v_y \cdot f)) = 1- \frac{1}{\pi}arccos(v_x\cdot v_y)
\end{equation*}
\begin{proof}
\begin{align*}
    p(sgn(v_x\cdot f) = sgn(v_y\cdot f)) &= 1 - p(sgn(v_x\cdot f) \ne sgn(v_y\cdot f))\\
    &= 1 - (p(v_x\cdot f\ge 0,  v_y\cdot f<0)+ p(v_y\cdot f\ge 0,  v_x\cdot f<0))\\
     &= 1 - 2*p(v_x\cdot f\ge 0,  v_y\cdot f<0) \quad \text{(By symmetry)}
\end{align*}

The set $\{f: v_x\cdot f\ge0, v_y\cdot f<0\}$ represents the intersection of two half spaces whose dihedral angle is $arccos(v_x\cdot v_y)$.
Since a full sphere (i.e., covers all possible random vectors) has a dihedral angle of $2\pi$, $p(v_x\cdot f\ge 0,  v_y\cdot f<0) = \frac{arccos(v_x\cdot v_y)}{2\pi}$. 
\begin{align*}
    p(sgn(v_x\cdot f) = sgn(v_y\cdot f)) &= 1 - \frac{arccos(v_x\cdot v_y)}{\pi}
\end{align*}
The lemma follows.
\end{proof}
\end{lemma}

By substituting from Lemma~\ref{lemma:1_app} to Equation~\ref{eq:omba_arm_app_1}, 
\begin{align}
    p(x \Rightarrow y)_{omba} &= 1-(1 - (1- \frac{arccos(v_x\cdot v_y)}{\pi})^{|F|})^{|H|}
    \label{eq:omba_arm_app}
\end{align}

\textbf{(2) Based on \textsc{Lift}.} In our negative sampling-based softmax approximation (see Section 4.2 in~\cite{anonymous2020omba}), we sample $|N_z|$ negative samples for every genuine product $x$ that co-occurred with $y$. Define the \textit{Lift} score between $x$ and $y$ in the empirical distribution of the training dataset and the noise distribution as $Lift(y,x)_{train}$ and $Lift(y,x)_{noise}$ respectively. Then, the combined \textit{Lift} score from both these distributions can be computed as a mixture of the both scores, which are weighted based on the number of samples coming from each:
\begin{equation}
    Lift(y,x) = \frac{1}{|N_z|+1}Lift(y,x)_{train} +  \frac{|N_z|}{|N_z|+1}Lift(y,x)_{noise} 
\end{equation}
However, the samples coming from $Lift(y,x)_{train}$ alone represent the true co-occurances of $x$ and $y$. Thus, the likelihood to have a strong association between product $x$ and $y$ (based on \textit{Lift}) can be computed as: 
\begin{align}
    p(x \Rightarrow y)_{lift} &= \frac{\frac{1}{|N_z|+1}Lift(y,x)_{train}}{\frac{1}{|N_z|+1}Lift(y,x)_{train} + \frac{|N_z|}{|N_z|+1}*Lift(y,x)_{noise}}\notag\\
    &= \frac{Lift(y,x)_{train}}{Lift(y,x)_{train} + |N_z|*Lift(y,x)_{noise}}
    \label{eq:lift_2}
\end{align}
By substituting $Lift(x,y)$ as $P(y|x)/P(y)$,
\begin{align}
    p(x \Rightarrow y)_{lift} &= 
    \frac{P_{train}(y|x)/P_{train}(y)}{P_{train}(y|x)/P_{train}(y) + |N_z|*P_{noise}(y|x)/P_{train}(y)}\notag\\
     &= 
    \frac{P_{train}(y|x)}{P_{train}(y|x) + |N_z|P_{noise}(y|x)}
    \label{eq:lift_2.1}
\end{align}
where $P_{train}$ and $P_{noise}$ represents the probabilities computed using the empirical distribution of the training dataset and the noise distribution respectively. In Equation~\ref{eq:lift_2.1}, $Lift_{noise}(x,y)$ is computed as $P_{noise}(y|x)/P_{train}(y)$, because the context products (product $y$ in this case) are always selected from the distribution of the training dataset (see Section 4.2 in~\cite{anonymous2020omba}).

According to negative sampling based softmax approximation in~\cite{mikolov2013distributed} (which is used to approximate softmax in this work), OMBA learns product representations that satisfy the following two constraints (see~\cite{mikolov2013distributed} for more details): 

\begin{align}
    |N_z|P_{noise}(y|x) &= |N_z|P_{noise}(y) = 1\label{eq:softmax_cons_1}\\
    \sum_{\forall x} exp(A*v_y \cdot v_x) &= 1 \text{  ($self$ $normalizing$ $constraint$)}\label{eq:softmax_cons_2}
\end{align}
where $v_x$ and $v_y$ denote the normalized embeddings in Equation~\ref{eq:omba_arm_app}. However, $v_x$  and $v_y$ are not necessarily normalized in Equation~\ref{eq:softmax_cons_2}. Thus, the constant $A$ is used to compensate for that. By substituting Equations~\ref{eq:softmax_cons_1} and~\ref{eq:softmax_cons_2} into Equation~\ref{eq:lift_2.1},
\begin{align}
    p(x \Rightarrow y)_{lift} 
     &= 
    \frac{P_{train}(y|x)}{P_{train}(y|x) + 1}&\text{  ($from$ $Eq.$ $~\ref{eq:softmax_cons_1}$)}\notag\\
    &= \frac{\frac{exp(A*v_x\cdot v_y)}{\sum_{\forall x} exp(A*v_y \cdot v_x)}}{\frac{exp(A*v_x\cdot v_y)}{\sum_{\forall x} exp(A*v_y \cdot v_x)} + 1} &\text{  ($from$ $Eq.$ $1$ $in~\cite{anonymous2020omba}$ )}\notag\\
    &= \frac{exp(A*v_x\cdot v_y)}{exp(A*v_x\cdot v_y) + 1}&\text{  ($from$ $Eq.$ $~\ref{eq:softmax_cons_2}$)}\notag\\
    &= \sigma(A*v_x\cdot v_y)\label{eq:lift_app}
\end{align}

\begin{figure*}[t]
\scriptsize
\centering
\begin{tikzpicture}
  \begin{axis}[width = 0.7\linewidth, 
  height= 4.5cm, domain = -1:1, samples = 500,
  legend style={at={(0.55,0.2)},anchor=west},
  xlabel = $v_x.v_y$,]
    \addplot[color = black, dashed]  {1/(1+exp(-4.3*x))};
    \addplot[color = gray] {1-(1-(1-acos(x)/180)^4)^11};
    \legend{$p_{omba}(x \Rightarrow y)$,$p_{lift}(x \Rightarrow y)$}
  \end{axis}
\end{tikzpicture}
\caption{Probability to have a strong association between $x$ and $y$ (as modelled by \textsc{OMBA} and \textsc{Lift}) with respect to the dot product of their embeddings for $|F|=4$, $|H|=11$, and $A=4.3$. The corresponding curves for other parameter values are simulated at~\href{https://www.desmos.com/calculator/tfssy4w7lh}{https://www.desmos.com/calculator/tfssy4w7lh}} \label{fig:optimal_values}
\vspace{-4mm}
\end{figure*}

\textbf{Optimal solution to $|F|$ and $|H|$.} Then, the integer solutions for parameters $|F|=4$ and $|H|=11$, and real solutions for $A=4.3$ are found such that $p(x \Rightarrow y)_{omba} = p(x \Rightarrow y)_{lift}$ (see Figure~\ref{fig:optimal_values} for the curves of $p(x \Rightarrow y)_{omba}$ and $p(x \Rightarrow y)_{lift}$ with the selected parameters). Such a selection of hyper-parameters theoretically guarantees that the rules produced from \textsc{OMBA-ARM} are higher in \textit{Lift}, which is a statistically well defined measure for strong associations.

\begin{figure*}[t]
\scriptsize
\centering
\subfloat[Effect of $d$]{%
\begin{tikzpicture}
\begin{axis}[
    xlabel={$d/100$},
    ylabel={MRR},
    xmin=0, xmax=5.5,
    ymin=0.45, ymax=0.65,
    xtick={0.5, 1, 2, 3, 4, 5},
    ytick={0.5,0.55, 0.6},
    legend pos=south east,
    ymajorgrids=true,
    grid style=dashed,
]
\addplot[
    color=black,
    mark=square*,
    ]
    coordinates {
    (0.5,0.4992)(1,0.5321)(2,0.58872)(3,0.6011)(4,0.6013)(5,0.6012)
    };
\end{axis}
\end{tikzpicture}
}\hspace{0.5em}
\subfloat[Effect of $|N_z|$]{%
\begin{tikzpicture}
\begin{axis}[
    xlabel={$|N_z|$},
    ylabel={MRR},
    xmin=0, xmax=6,
    ymin=0.595, ymax=0.605,
    xtick={1, 2, 3, 4 , 5},
    ytick={0.59,0.6,0.61},
    legend pos=south east,
    ymajorgrids=true,
    grid style=dashed,
]
\addplot[
    color=black,
    mark=square*,
    ]
    coordinates {
    (1,0.5986)(2,0.6002)(3,0.6013)(4,0.6011)(5,0.6015)
    };
\end{axis}
\end{tikzpicture}%
}\hspace{0.5em}
\subfloat[Effect of $\tau$]{%
\begin{tikzpicture}
\begin{axis}[
    xlabel={$\log(\tau)$},
    ylabel={MRR},
    xmin=-4, xmax=2,
    ymin=0.59, ymax=0.61,
    xtick={ -3,-2,-1,0,1},
    ytick={0.59,0.6,0.61},
    legend pos=south east,
    ymajorgrids=true,
    grid style=dashed,
]
\addplot[
    color=black,
    mark=square*,
    ]
    coordinates {
    (-3,0.5934)(-2,0.5976)(-1,0.6011)(0,0.6013)(1,0.6001)
    };
\end{axis}
\end{tikzpicture}
}
\vspace{-1mm}
\caption{MRRs for the \textit{intra-basket item retrieval} task with different hyper-parameter settings using CJ dataset}\label{fig:para_sens}
\vspace{-2mm}
\end{figure*}

\begin{figure*}[t]
\scriptsize
\centering
\subfloat[CJ Dataset]{%
\begin{tikzpicture}
\begin{axis}[
    width=4.3cm,
    xlabel={$k$},
    ylabel={R@k},
    xmin=0, xmax=6,
    ymin=0.2, ymax=1,
    xtick={1, 2, 3, 4, 5},
    ytick={0.25,0.5,0.75},
    legend style={at={(0.5,-0.2)},anchor=north,legend columns=1},
    ymajorgrids=true,
    grid style=dashed,
]
\addplot[
    color=black,
    mark=*,
    ]
    coordinates {
    (1, 0.4326)(2,0.5969)(3,0.6967)(4,0.7724)(5,0.8328)
    };
\addplot[
    color=black!60,
    mark=square*,
    ]
    coordinates {
    (1, 0.3392)(2,0.4804)(3,0.5867)(4,0.6693)(5,0.7373)
    };
\addplot[
    color=gray,
    mark=triangle,
    ]
    coordinates {
    (1, 0.3776)(2,0.5063)(3,0.5996)(4,0.6668)(5,0.7251)
    };
\legend{\textsc{OMBA}, \textsc{Triple2Vec}, \textsc{Lift}}    
\end{axis}
\end{tikzpicture}
}\hspace{0.5em}
\subfloat[IC Dataset]{%
\begin{tikzpicture}
\begin{axis}[
    width=4.3cm,
    xlabel={$k$},
    ylabel={R@k},
    xmin=0, xmax=6,
    ymin=0.2, ymax=1,
    xtick={1, 2, 3, 4, 5},
    ytick={0.25,0.5,0.75},
    legend style={at={(0.5,-0.2)},anchor=north,legend columns=1},
    ymajorgrids=true,
    grid style=dashed,
]
\addplot[
    color=black,
    mark=*,
    ]
    coordinates {
    (1, 0.5859)(2,0.7760)(3,0.8690)(4,0.9205)(5,0.9506)
    };
\addplot[
    color=black!60,
    mark=square*,
    ]
    coordinates {
    (1, 0.4348)(2,0.6208)(3,0.7418)(4,0.8254)(5,0.8849)
    };
\addplot[
    color=gray,
    mark=triangle,
    ]
    coordinates {
    (1, 0.2655)(2,0.3612)(3,0.5446)(4,0.7681)(5,0.8868)
    };
\legend{\textsc{OMBA}, \textsc{Triple2Vec}, \textsc{Lift}}    
\end{axis}
\end{tikzpicture}%
}\hspace{0.5em}
\subfloat[TF Dataset]{%
\begin{tikzpicture}
\begin{axis}[
    width=4.3cm,
    xlabel={$k$},
    ylabel={R@k},
    xmin=0, xmax=6,
    ymin=0.2, ymax=0.8,
    xtick={1, 2, 3, 4, 5},
    ytick={0.25,0.5,0.75},
    legend style={at={(0.5,-0.2)},anchor=north,legend columns=1},
    ymajorgrids=true,
    grid style=dashed,
]
\addplot[
    color=black,
    mark=*,
    ]
    coordinates {
    (1, 0.3466)(2,0.4818)(3,0.5844)(4,0.6657)(5,0.7365)
    };
\addplot[
    color=black!60,
    mark=square*,
    ]
    coordinates {
    (1, 0.3040)(2,0.4342)(3,0.5361)(4,0.6194)(5,0.6927)
    };
\addplot[
    color=gray,
    mark=triangle,
    ]
    coordinates {
    (1, 0.2981)(2,0.4169)(3,0.5024)(4,0.5691)(5,0.6269)
    };
\legend{\textsc{OMBA}, \textsc{Triple2Vec}, \textsc{Lift}}    
\end{axis}
\end{tikzpicture}
}
\caption{Recall at different k values for \textit{intra-basket item retrieval}}\label{fig:diff_recall_k}
\vspace{-2mm}
\end{figure*}

\section{Hyper-parameter Sensitivity of OMBA}\label{sec:4}
Figure~\ref{fig:para_sens} shows the sensitivity of the hyper-parameters of the OMBA-OME module, which returns the ideal parameters for the \textit{intra-basket item retrieval task} as: (1) $d=300$; (2) $|N_z|=3$; and (3) $\tau=0.1$.

\section{More results on \textit{intra-basket item retrieval}}\label{sec:5}
Figure~\ref{fig:diff_recall_k} shows the results, evaluated using recall at different $k$ values, for the \textit{intra-basket item retrieval} task using the CJ, IC, and TF datasets. The results show that \textsc{OMBA} consistently outperforms other two baselines for different $k$ values.

\bibliographystyle{splncs04}
\bibliography{Supplement}